%% file: GPoolSGNNs_eusipco2020_v02.tex
\documentclass[conference]{IEEEtran}
\IEEEoverridecommandlockouts
\usepackage[english]{babel}
\usepackage{ifpdf}
\usepackage{cite} 
\usepackage{url}
\usepackage[draft,bookmarks=false]{hyperref} 
\ifpdf
	\usepackage[pdftex]{graphicx}
	\graphicspath{{./figures/}}
\else
	\usepackage[dvips]{graphicx}
	\graphicspath{./figures/}
\fi
\usepackage{color}
\usepackage{pgf, tikz, pgfplots}
\usetikzlibrary{shapes, arrows, automata}
\usepackage{caption} 
\usepackage{subcaption} 
\usepackage{amsmath}
\usepackage{amsfonts, amssymb, amsthm}
\usepackage{mathrsfs}
\usepackage{upgreek}
\usepackage{algorithm,algpseudocode}
\usepackage{enumerate}
\usepackage{multirow}
\usepackage{rotating}
\input{mySections.tex}

\input{mySymbol.sty}

\newtheorem{theorem}{\hspace{0pt}\bf Theorem}

\newcommand{\QED}{\hfill\ensuremath{\blacksquare}}

\title{Graphon Pooling in Graph Neural Networks}
\author{Alejandro~Parada-Mayorga, Luana~Ruiz~
        and~Alejandro~Ribeiro
\thanks{All the authors contributed equally. The authors are with the Dept. of Electrical and Systems Eng., Univ. of Pennsylvania. Email: \{alejopm,rubruiz,aribeiro\}@seas.upenn.edu.
Supported by NSF CCF 1717120, ARO W911NF1710438, ARL DCIST CRA W911NF-17-2-0181, ISTC-WAS and Intel DevCloud.
}
}

\begin{document}
%
\maketitle
\begin{abstract}
Graph neural networks (GNNs) have been used effectively in different applications involving the processing of signals on irregular structures modeled by graphs. Relying on the use of shift-invariant graph filters, GNNs extend the operation of convolution to graphs. However, the operations of pooling and sampling are still not clearly defined and the approaches proposed in the literature either modify the graph structure in a way that does not preserve its spectral properties, or require defining a policy for selecting which nodes to keep. In this work, we propose a new strategy for pooling and sampling on GNNs using graphons which preserves the spectral properties of the graph. To do so, we consider the graph layers in a GNN as elements of a sequence of graphs that converge to a graphon. In this way we have no ambiguity in the node labeling when mapping signals from one layer to the other and a spectral representation that is consistent throughout the layers. We evaluate this strategy in a synthetic and a real-world numerical experiment where we show that graphon pooling GNNs are less prone to overfitting and improve upon other pooling techniques, especially when the dimensionality reduction ratios between layers is large. 
\end{abstract}
%

\section{Introduction}\label{sec:intro}
\input{sec_intro.tex}

\section{Convolutional Graph Neural Networks}\label{sec:gnn}
\input{sec_gnn}

\input{sec_graphon_nn_v02}

\section{Numerical Experiments} \label{sec:sims}
\input{sec_sims}

\section{Conclusions} \label{sec:conclusions}
\input{sec_conclusions}

\bibliographystyle{IEEEbib}
\bibliography{myIEEEabrv,biblioEusipcoGRNN,bib-graphon}

\end{document}

%% file: mySections.tex
\usepackage{needspace}



%% file: sec_intro.tex


Convolutional neural networks (CNNs) have become essential in diverse applications, overcoming the scalability issues of traditional neural networks related to the number of learnable parameters of the architecture, which previously led to computational issues. In a CNN, the linear transformation is regularized using convolutions between the signals and a set of filters whose parameters can be chosen to be small and independent of the size of the data. This allows the architecture to scale, and plenty of numerical evidence shows impressive performance in classification and regression applications~\cite{bronstein17-geomdeep,bruna14-deepspectralnetworks}. On the other hand, CNNs can only process information defined on Euclidean spaces, which makes them unsuitable for handling data on irregular domains, i.e., where the convolution operation is not trivially defined. Graph neural networks (GNNs) emerged as tools to generalize CNNs to non-Euclidean domains by leveraging consolidated developments of graph signal processing and exploiting the shift operator to perform graph convolutions~\cite{gama18-gnnarchit,ruiz18-local}.

In this paper, we introduce graphons as a way to capture structural properties of the graphs in the layers of a GNN and to preserve these properties through the pooling operation. The approach we propose exploits the spectral similarities of the graphs that are obtained from a graphon. This allows reducing the losses and distortions introduced by coarsening or zero-padding. Additionally, our approach provides a natural ordering of the nodes that allows to map the downsampled signal from one layer to the next without ambiguity. This eliminates the node selection step and thus offers a method that is computationally more efficient than coarsening approaches presented in the literature. This approach also opens the door for a new class of GNNs where the graphs in each layer are known with some degree of uncertainty as they are obtained as samples of a probability space defined by a graphon.

Besides the extension of the convolution operation to GNNs, the notion of pooling plays a central role as it provides dimensionality reduction from one layer to the other. In CNNs this step is straightforward as the reduction of the dimension is performed on a Euclidean space, so the  intrinsic properties of the domain of the signal remain invariant. In GNNs, that is not the case. In~\cite{defferrard17-cnngraphs}, graph coarsening is considered as a way to provide dimensionality reduction, however as pointed out in~\cite{gama18-gnnarchit} this process can distort the structure and properties of the original graph, leading to loss in performance. In~\cite{gama18-gnnarchit}, zero-padding is proposed to replace the sampling operation between layers. This approach is the functional equivalent of doing coarsening with the induced subgraphs that are obtained from the nodes whose signal value is different than zero after zero-padding. The downside of this method is that it requires an optimal selection of the nodes to keep which usually relies in sampling techniques like the ones presented in~\cite{anisortegaproxis}.  

We perform numerical simulations in two scenarios where GNNs employing different pooling strategies are used to highlight the advantages of graphon pooling.
In the first, we consider the problem of predicting the source node of a synthetic diffusion process on a graph. In the second, we use the MovieLens 100k dataset to construct a user similarity network and predict the ratings a user has given to a set of movies. Graphon coarsening can be seen to outperform other pooling techniques in terms of accuracy and mean squared error in both cases, especially when the dimensionality reduction ratio between layers of the GNN is big. We also observe distinctively less overfitting in comparison with other approaches. 

This paper is organized as follows. Section II provides basic concepts about GNNs. In Section III we introduce the basics of graphon signal processing and in Section IV we discuss graphon neural networks and graphon-based pooling strategies. Section V presents a set of numerical experiments and we finish with conclusions in Section VI.

%% file: sec_gnn.tex


We define graphs as triplets $G=(V(G),E(G),\alpha)$, with set of nodes $V(G)=\{1,\ldots,N\}$, set of edges $E(G)$ and weight function $\alpha: E(G)\rightarrow\mathbb{R}^{+}$. Graph signals are functions $\mathbf{x}: V(G)\rightarrow\mathbb{R}$. The image of $\mathbf{x}(V(G))$ is identified with vectors in $\mathbb{R}^{N}$ and the value of $\mathbf{x}$ at a given node $i\in V(G)$ is given by the component $[\mathbf{x}]_{i}$. We focus on undirected graphs where the elements of $E(G)$ are considered to be unordered pairs and therefore $\alpha(\{i,j\})=\alpha(\{j,i\})$. The sparsity pattern of $G$ is captured by its adjacency matrix $\mathbf{W}$ whose elements are given by $[\mathbf{W}]_{ij}=\alpha(\{i,j\})$, where $\{i,j\}\in E(G)$. 

The building block in GSP is the graph shift operator (GSO) of $G$, defined as any matrix $\mathbf{S}\in\mathbb{R}^{N\times N}$ operating on signals defined on $V(G)$ that satisfies that $[\mathbf{S}]_{ij}=0$ if $i\neq j$ and $\{i,j\}\notin E(G)$. A particular case of this operator is $\mathbf{S}=\mathbf{W}$ which is the choice for this paper, but other operators like the Laplacian can be also considered~\cite{gama18-gnnarchit}. With the specification of $\mathbf{S}$ we can define \textit{graph convolutions} \cite{gama18-gnnarchit} as
\begin{equation}\label{eq:Hingraph}
\mathbf{H}\mathbf{x}=\sum_{k=0}^{K-1}h_{k}\mathbf{S}^{k}\mathbf{x}
\end{equation}
where the $h_{k}$ are the graph filter taps. Like time convolutions, graph convolutions are also shift-invariant: it is easy to see that a shift to the input $\bbS\bbx$ produces the shifted output $\bbS\bbH\bbx$.

For a training set $\mathcal{T}=\{(\mathbf{x},\mathbf{y})\}$ with input samples $\mathbf{x}$ and output samples $\mathbf{y}$, a GNN learns a representation or mapping capable of producing output estimates to unseen inputs $\hat{\mathbf{x}}\notin\mathcal{T}$. The GNN is a stacked architecture consisting of graph layers where information is mapped from one graph layer to the other by means of a linear operator and a point-wise nonlinearity~\cite{gama18-gnnarchit}. The output of the $\ell^{th}$ layer is given by
\begin{equation}\label{eq:Hingnn}
\mathbf{x}_{\ell}=\sigma_{\ell}\left( \mathbf{H}_{\ell}\mathbf{x}_{\ell-1}\right)
\end{equation}
where $\mathbf{x}_{\ell-1}$ is the output of the layer $\ell-1$, $\mathbf{H}_{\ell}$ is a graph convolution like in \eqref{eq:Hingraph} and $\sigma_{\ell}$ is the composition of a point-wise nonlinear operator and a pooling operator. 
The filter taps $h_{k\ell}$ are calculated in the training of the GNN and $K\ll N$. Like in the case of CNNs, several versions of the operator in~(\ref{eq:Hingraph}) can be used in the same layer of a GNN considering different values of the $h_{k\ell}$ to extract multiple features.

The notion of pooling in the GNN is described by the properties of the pooling operation in $\sigma_{\ell}$ which, like in CNNs, composes a summarizing operator like the average or the max with a sampling operator. One substantial difference with respect to CNNs is that in the sampling
operation, the assignment of the sample at a given node in layer $\ell-1$ to a node in the next layer $\ell$ is not trivial. Indeed, there are essentially two issues regarding to the pooling operation on GNNs. The first is determining the graph in each layer taking into account that the dimensionality of the signals is being reduced. The second is the mapping of the sampled values of a signal to the nodes of the graph of the next layer. As we will see in the following sections, graphons provide a natural way to tackle these issues.

%% file: sec_graphon_nn_v02.tex

\section{Graphon Signal Processing} \label{sec:graphon}
A graphon is a bounded symmetric measurable function $W:[0,1]^2 \to [0,1]$. Originally conceived as ``limit graphs'' of sequences of dense graphs whose number of nodes grows to infinity~\cite{lovasz2012large}, graphons have also been used as generative models for random graphs~\cite{avella2018centrality}. In these graphs, the entries of the adjacency matrix are sampled as Bernoulli random variables with probability $\kappa W(u_{i},u_{j})$, where the constant $\kappa$ defines the sparsity of the graph and $u_{i},u_{j}\in [0,1]$ are selected at random. 

Thinking of the graphon as a graph with an uncountable number of nodes, we can represent data on graphons as graphon signals $X: [0,1] \to \reals$, which map points of the unit interval (the graphon ``nodes'') to the real line~\cite{ruiz2019graphon}. In the same way that graphs can be obtained from graphons by evaluating $W(u,v)$ at points $u_i \in [0,1]$, a graph signal $\bbx$ can be obtained from $X$ by setting $[\bbx]_i = X(u_i)$.


Conversely, we can also define graphons and graphon signals induced by graphs and graph signals. For any finite graph $G$ with $\vert V(G)\vert=N$ and shift operator $\mathbf{S}$, we can associate a graphon $W_{G}$ given by
\begin{equation}
W_{G}(u,v)=\sum_{i,j=1}^{N}[\mathbf{S}]_{ij}\chi_{\tau_{i}}(u)\chi_{\tau_{j}}(v)
\end{equation}
where $\chi_{\tau_{i}}(u)$ is the characteristic function of the interval $\tau_{i}$ where $\{\tau_{i}\}_{i=1}^{N}$ forms a partition of $[0,1]$. Accordingly, if there is a graph signal $\mathbf{x}$ defined on $G$ it is possible to associate to it a step-function graphon signal given by
\begin{equation}
X_{G}(u)=\sum_{i=1}^{N}[\mathbf{x}]_{i}\chi_{\tau_{i}}(u).
\end{equation}

Every graphon $W$ induces a Hilbert-Schmidt operator acting on graphon signals $X$ as
\begin{equation}
\left(T_{W}X\right)(v)=\int_{0}^{1}W(u,v)X(u)du .
\end{equation}
We refer to $T_{W}$ as the \textit{graphon shift operator} (WSO). This notion of shift allows extending filter convolutions to graphon signals, and so we define shift-invariant graphon filters as $H=\sum_{k=0}^{K}h_{k}T_{W}^{k}$ where $T_{W}^{k}$ stands for $k$ consecutive applications of $T_{W}$. Applied to $X$, this yields
\begin{equation} \label{eq:Hingraphon}
H\ast X=\sum_{k=0}^{K}h_{k}T_{W}^{k}X.
\end{equation} 
Note that \eqref{eq:Hingraphon} is the continuous counterpart of \eqref{eq:Hingraph} and that, in particular, for graphons and graphon signals induced by graphs and graph signals these expressions yield the same result. This parallel is important because it allows thinking of graph filtering operations as graphon filtering operations. The advantage of operating in the graphon domain is that we can leverage the probabilistic interpretation of graphons to sample graphs of smaller or larger size whose structure is faithful to that of the original graph. This motivates the definition of graphon pooling in Section \ref{sec:graphon_nn}.

\section{Graphon Pooling}\label{sec:graphon_nn}
\begin{figure}
 \centering\includegraphics[angle=-90,scale=0.42]{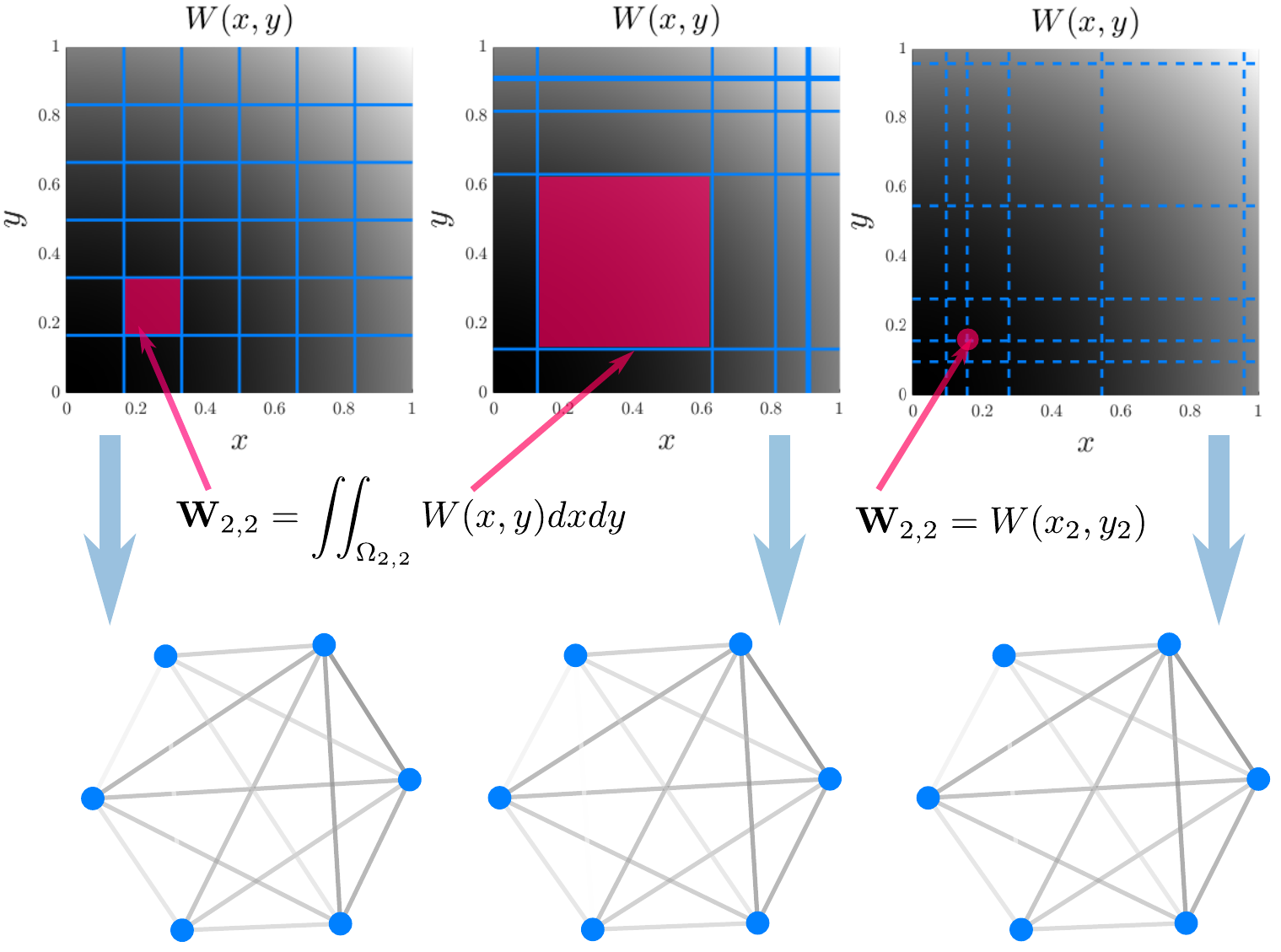}
 \caption{Different ways to obtain a graph from a graphon.}
 \label{fig:graphfromgraphon}
\end{figure}

We propose to process information on the GNN using the underlying graphon model of the graph to generate the graphs at each layer of the GNN. This provides a natural way of performing pooling, as dimensionality reduction can be achieved by sampling graphs of smaller size.

Finite graphs can be obtained from graphons in multiple ways. For instance, as described in Section \ref{sec:graphon} and depicted in the third diagram of Figure \ref{fig:graphfromgraphon}, an $N$-node graph $G=(V(G),E(G),\alpha)$ can be \textit{sampled} from the graphon by using a function $\rho: V(G)\rightarrow [0,1]$ to map each node $i\in V(G)$ to a number $\rho(i)\in [0,1]$, such that, for any two nodes $i,j\in V(G)$, we have $\{i,j\}\in E(G)$ and $\alpha(\{i,j\})=W(\rho(i),\rho(j))$. Another possibility is to generate graphs through \textit{piece-wise approximations}, as pictured in the two left diagrams of Figure \ref{fig:graphfromgraphon}. In this case, the weight of the edge $(i,j)$ is given by
\begin{equation} \label{eqn:integration}
\alpha(\{i,j\})=\frac{1}{\Delta}\int_{\rho(j)}^{\rho(j+1)}\int_{\rho(i)}^{\rho(i+1)}W(u,v)dudv
\end{equation}
for $1<i,j<N$, where $\Delta=(\rho(i+1)-\rho(i))(\rho(j+1)-\rho(j))$ and $\rho(i)\leq\rho(i+1)$ for all $i$.

\begin{figure}
 \centering\includegraphics[angle=-90,scale=0.53]{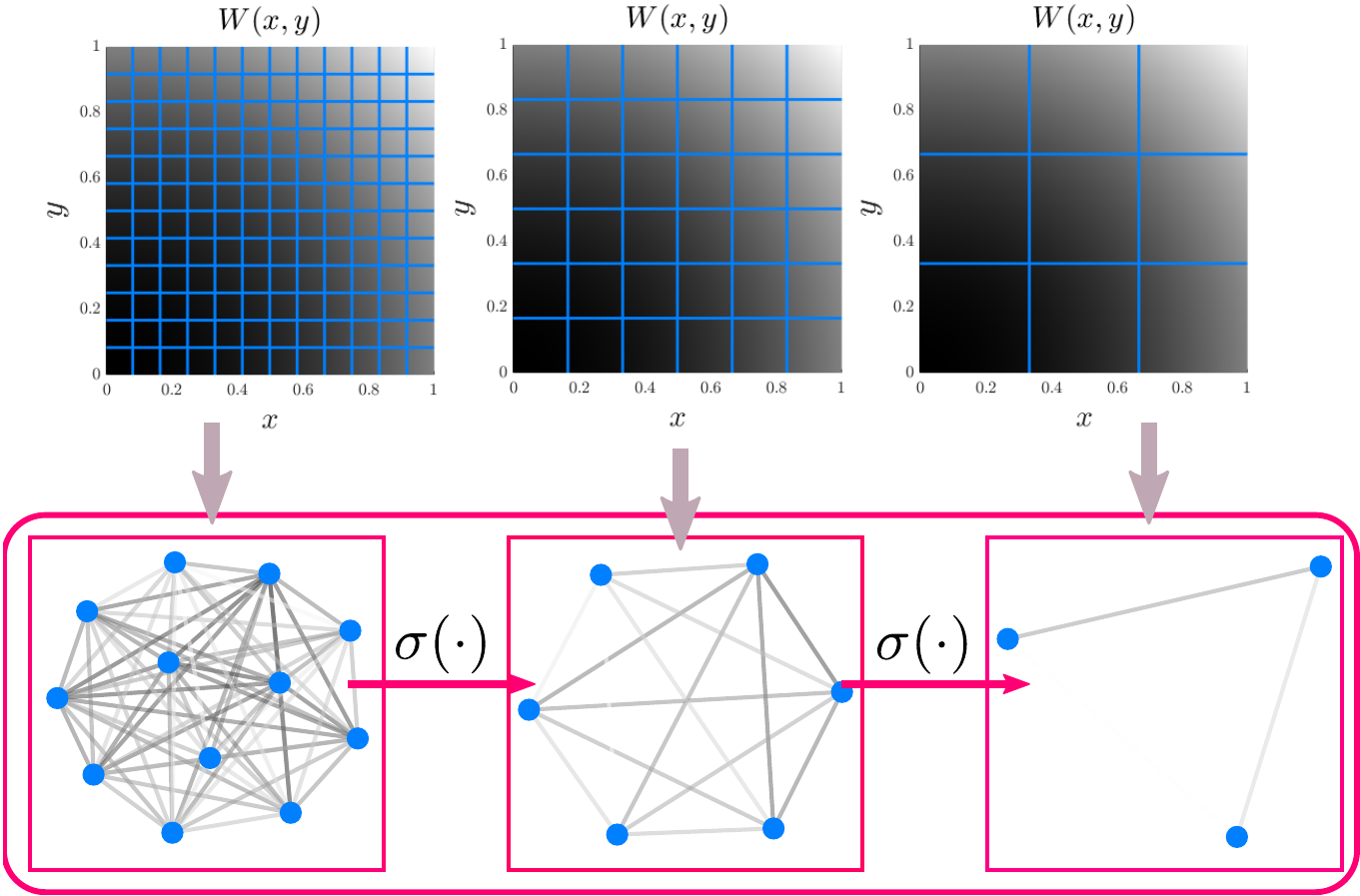}
 \caption{Graphon Neural Network with three layers}
 \label{fig:graphon_nn}
\end{figure}

We focus on the piece-wise approximation method, and generate the graphs in each layer of the network using~(\ref{eqn:integration}) to obtain the $\ell$th layer GSO $\bbS_\ell \in \reals^{N_\ell \times N_\ell}$. The operations performed in each layer of this \textit{graphon neural network} are then given by~(\ref{eq:Hingnn}), where $\bbH_\ell = \sum_{k=0}^{K-1} h_{k\ell}\bbS_\ell^k$ and $N_\ell \leq N_{\ell-1}$ for $1 \leq \ell < L$. When $N_\ell < N_{\ell-1}$, the convolution output $\bbH_{\ell-1}\bbx_{\ell-1} \in \reals^{N_{\ell-1}\times N_{\ell-1}}$ at each layer is mapped to $\bbx_\ell \in \reals^{N_\ell}$ using the natural ordering of the nodes induced by the domain of the graphon signal (i.e, the unit interval). 
This pooling method is illustrated in Fig.~(\ref{fig:graphon_nn}), where a GNN with three layers is built from a graphon. 
\subsection{Spectral motivation}
We recall that for $\mathbf{x}$ defined on $G$, the graph Fourier transform (GFT) is obtained by diagonalizing the GSO as $\mathbf{S}=\mathbf{U}\mathbf{\Lambda}\mathbf{U}^{T}$ and calculating $\hat{\mathbf{x}}=\mathbf{U}^{T}\mathbf{x}$, where $\mathbf{\Lambda}$ has $N$ eigenvalues given by $\sigma_{1}\geq\sigma_{2}\geq\cdots 0$ and $\sigma_{-1}\leq\sigma_{-2}\leq\cdots 0$. The component of $\hat{\mathbf{x}}$ associated with $\sigma_{j}$ is represented as $\hat{\mathbf{x}}(\sigma_{j})$.
As for the graphon, since $T_{W}$ is compact and bounded it possesses a countable set of eigenvalues given by $1\geq\sigma_{1}\geq\sigma_{2}\geq\cdots 0$ and $-1\leq\sigma_{-1}\leq\sigma_{-2}\leq\cdots 0$ (with $0$ as an accumulation point), and associated eigenfunctions $\varphi_{i}(u)$. Using the functions $\varphi_{i}$, it is possible to obtain a spectral decomposition of the graphon signal $X$ as
\begin{align}
X=\sum_{i=1}^{\infty}\left(\hat{X}(\sigma_{i})\varphi_{i}+\hat{X}(\sigma_{-i})\varphi_{-i}\right) \\
\mbox{where }\hat{X}(\sigma_{j})=\int_{0}^{1}X(u)\varphi_{j}(u)du.
\end{align}
The discrete function $\hat{X}$ is known as the \textit{graphon Fourier transform} (WFT) of $X$ and the graphon signal $X$ is called $\omega$-bandlimited if there exists $\omega\in (0,1)$ such that $\hat{X}(\sigma_{j})=0$ for all $\vert\sigma_{j}\vert<\omega$~\cite{ruiz2019graphon}.

The motivation for graphon pooling comes from the fact that graphs obtained from the same graphon with the same procedure have similar spectral characteristics. Indeed, as proven in~\cite{lovasz2012large} (Theorem 11.53), the eigenvalues of the graphs in a convergent graph sequence converge to the eigenvalues of the graphon itself, and, as proved in~\cite{ruiz2019graphon} and reproduced here as Theorem \ref{theorem:graphFTconv}, the graph Fourier transform converges to the \textit{graphon Fourier transform}. 

\begin{theorem}(\cite{ruiz2019graphon})\label{theorem:graphFTconv}
Let $\{(G_{n},\mathbf{x}_{n})\}$ be a sequence of graph signals converging to the $\omega-$bandlimited graphon signal $(W,X)$, where $T_{W}$ does not have repeated eigenvalues. Then, there exists a sequence of permutations $\pi_{n}$ such that GFT$(\pi_{n}(G_{n}),\pi_{n}(\mathbf{x}))\rightarrow$WFT$(W,X)$ in the sense that $[\hat{\mathbf{x}}]_{j}/\sqrt{n}\rightarrow\hat{X}(\sigma_{j})$ as $n\rightarrow\infty$.
\end{theorem}
An important implication of Theorem~\ref{theorem:graphFTconv} with respect to graphon pooling is that, if the signals considered in each layer are equivalent in spectrum---i.e., if they are all bandlimited with the same bandwidth---, then filtering on every graph obtained from the graphon is equivalent. We observe that this attribute is shared with the approach proposed in~\cite{gama18-gnnarchit}, but graphon pooling has the advantage of eliminating ambiguity when mapping signals from one layer to the other.

%% file: sec_sims.tex


%

\begin{figure*}[t]
	\centering
	\begin{subfigure}{.32\textwidth}
		\centering
		\includegraphics[width=\textwidth]{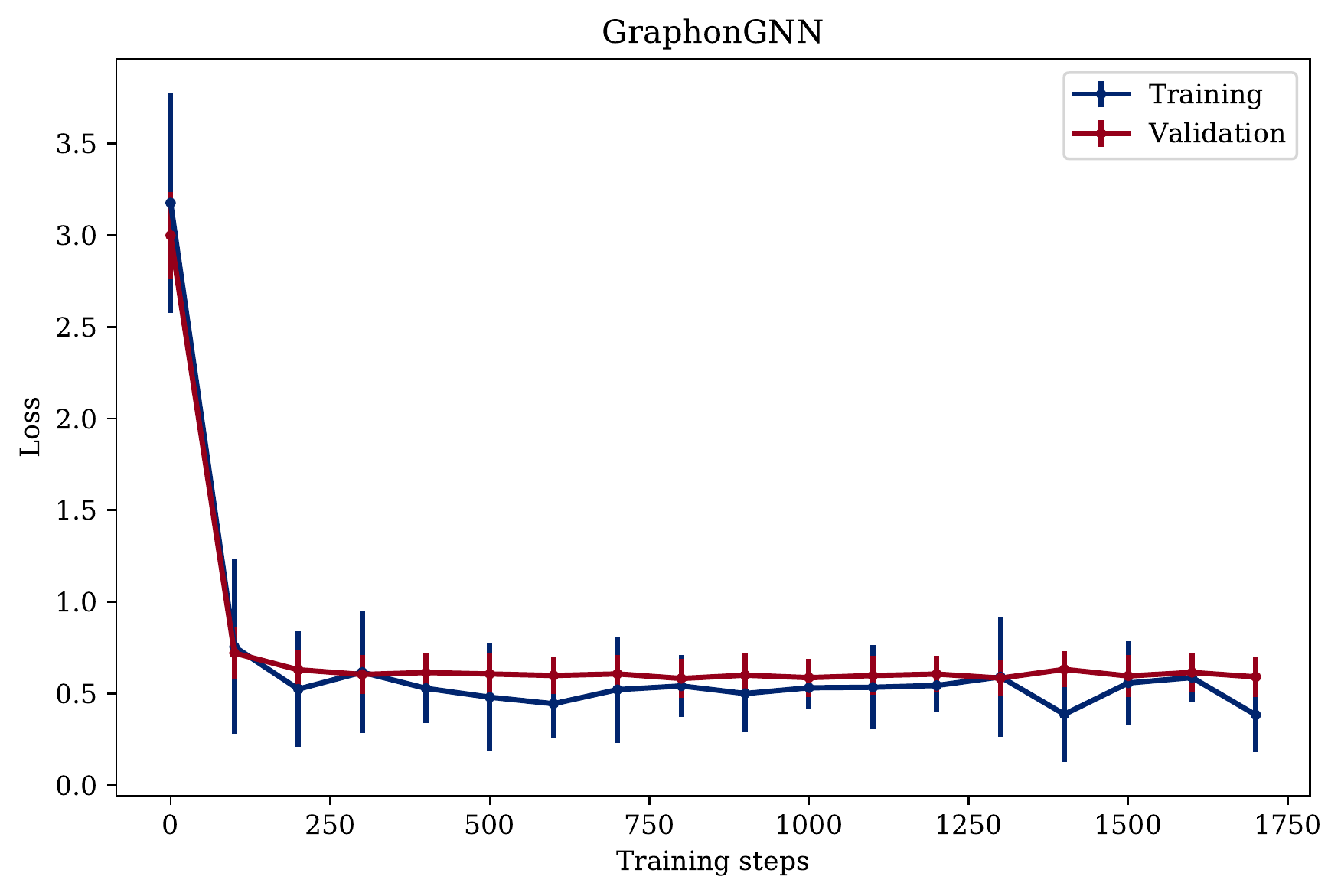}
		\caption{Graphon poooling}
		\label{graphon_training_loss}
	\end{subfigure}
	\hfill
	\begin{subfigure}{.32\textwidth}
		\centering
		\includegraphics[width=\textwidth]{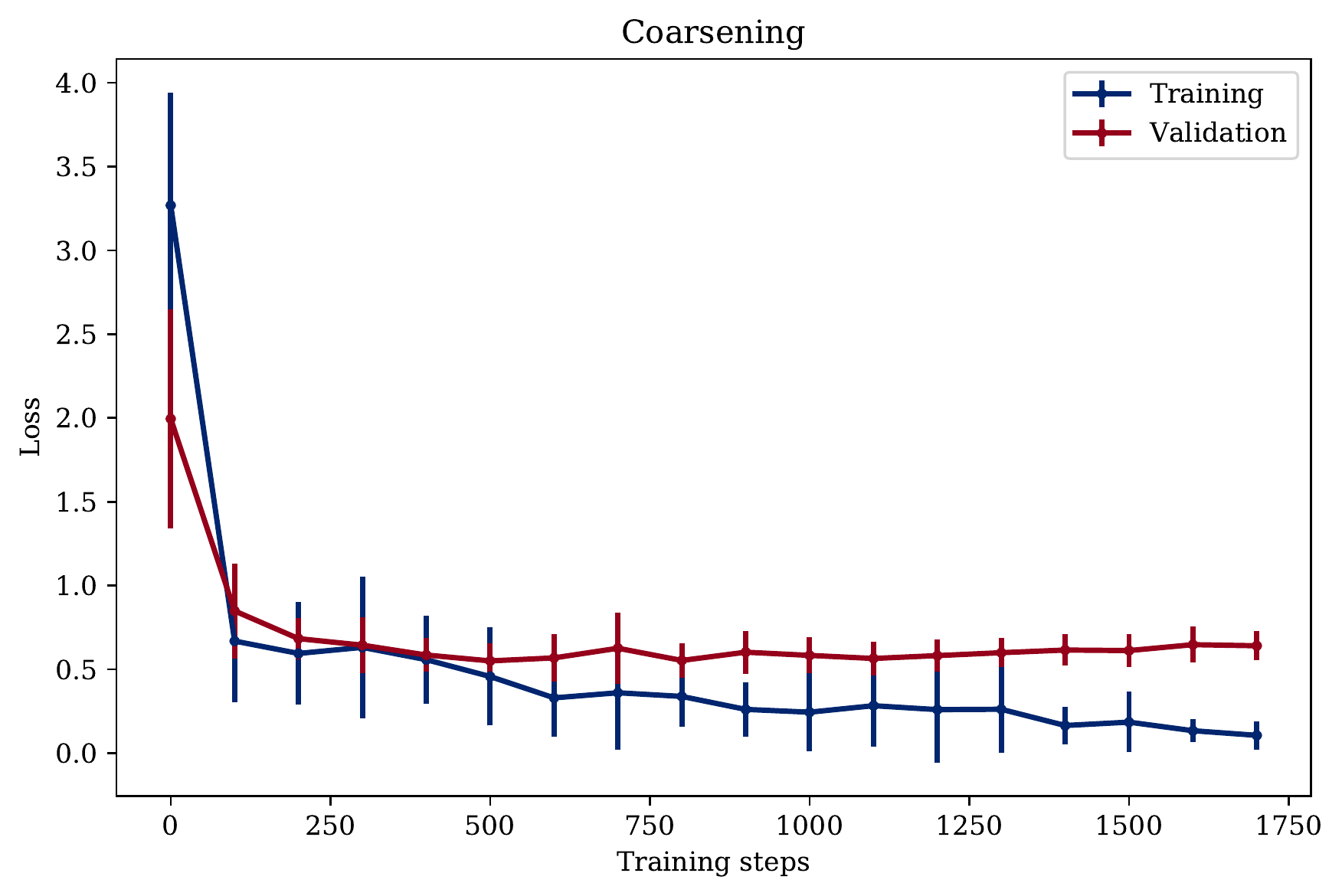} 
		\caption{Graph coarsening}
		\label{coarsening_training_loss}
	\end{subfigure}
	\hfill
	\begin{subfigure}{.32\textwidth}
		\centering
		\includegraphics[width=\textwidth]{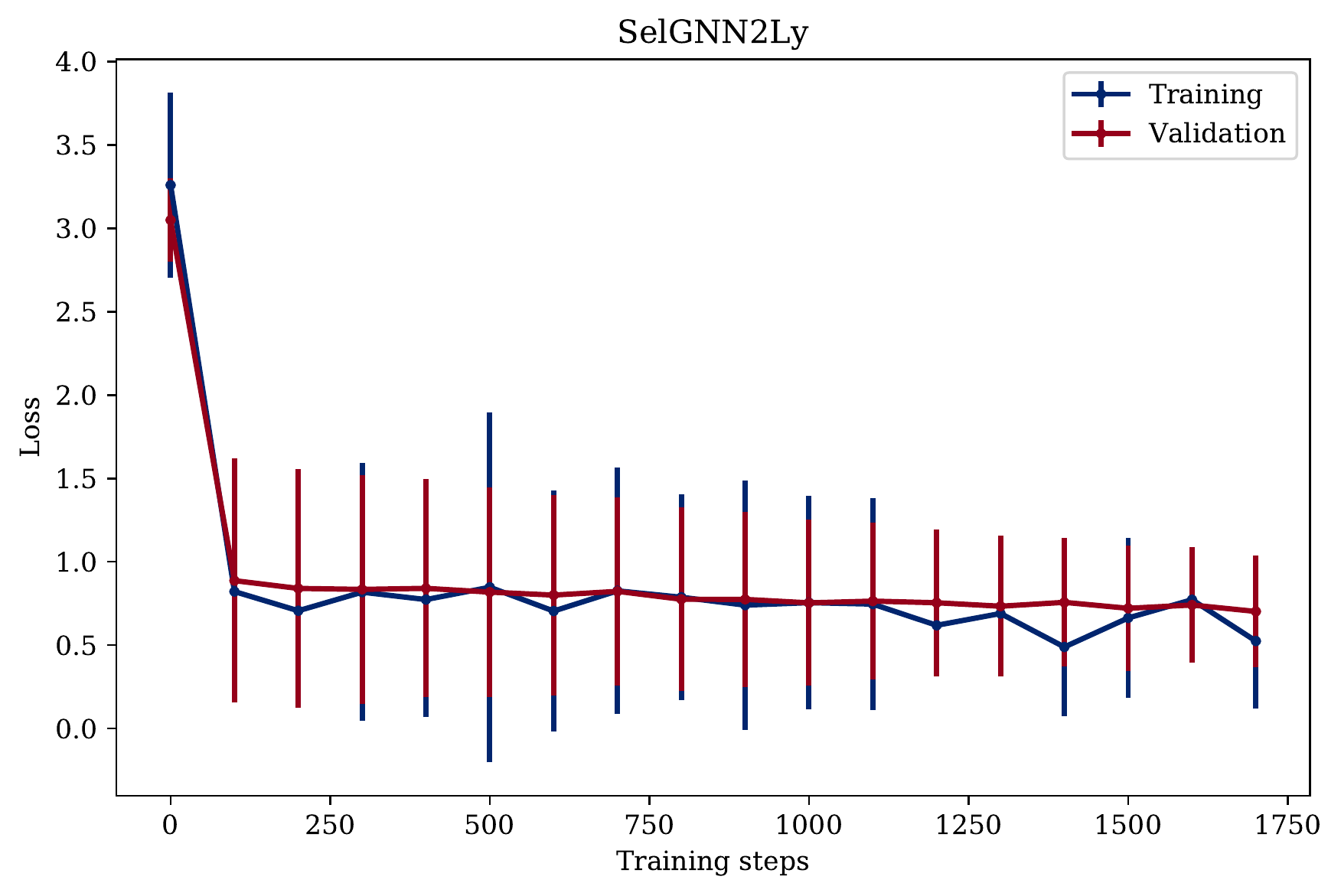} 
		\caption{Selection GNN}
		\label{selection_training_loss}
	\end{subfigure}
	\caption{Average training and validation losses for predicting the ratings given by user 1 over 10 different training-test splits.}
	\label{fig:loss}
\end{figure*}	



We evaluate the advantages of graphon pooling in two sets of experiments. The first is a classification problem where, given a synthetically generated graph diffusion process on graphs obtained from several graphon models, the objective is to predict the node corresponding to the process' seed. The second consists of predicting the rating given by a user to a movie from ratings that other users gave to the same movie. The users are connected through a user similarity network built from real data, which we use to build a piecewise constant graphon. 

In all experiments, we compare graphon pooling with two other pooling strategies: graph coarsening \cite{defferrard17-cnngraphs} and selection GNNs \cite{gama18-gnnarchit}. All of the GNNs are trained in parallel using the ADAM algorithm for stochastic optimization \cite{kingma17-adam} with decaying factors $\beta_1= 0.9$ and $\beta_2 = 0.999$.

\subsection{Source localization} \label{sbs:sourceloc}

Consider an $N$-node graph with shift operator $\bbS \in \reals^{N \times N}$ and let $\bbx_0 \in \reals^N$ be a graph signal such that $[\bbx_0]_i = 1$ for node $i = c$ and $0$ everywhere else. We define the graph diffusion process $\{\bbx_t\}$ as \cite{ruiz2020gated}
\begin{equation}
\bbx_t = \bbS^t \bbx_0
\end{equation}
and consider the problem of locating the source node $c$ given $\bbx_t$ for arbitrary $t$.
The node $c$ is chosen among $10$ possible sources, making for a classification problem on $C=10$ classes. 

We train GNNs to solve this problem by optimizing the cross-entropy loss on $1,000 \ (\bbx_t,c)$ training samples divided in batches of $20$, over $300$ epochs and with learning rate $5 \times 10^{-4}$. The learned models are validated and tested by evaluating the classification accuracy on sets containing $240$ and $200$ input-output samples respectively. All models consist of GNNs with $L=2$ layers, where both layers have $F_1 = F_2 = 8$ output features and graph convolutional filters with $K_1 = K_2 = 5$ filter taps each. 

First, we analyze the advantages of graphon pooling in graphs obtained from different graphon models. We consider three graphons: the exponential graphon $W(x,y) = \exp (-\beta(x-y)^2)$ with $\beta=2.3$; the bilinear graphon $W(x,y) = xy$; and the polynomial graphon $W(x,y) = 0.5(x^2+y^2)$. We generate graphs with $N=200$ nodes by integrating $W(x,y)$ as in equation \eqref{eqn:integration}, and the numbers of selected nodes in the first and second layers of all GNNs are $N_1 = 20$ and $N_2=10$ respectively. The test classification accuracies achieved with graphon pooling, graph coarsening and the selection GNN are presented in Table \ref{tb:graphons} for all three graphons. We report the average accuracy and standard deviation for models trained on 10 different dataset realizations. We observe that, in the case of the exponential and bilinear graphons, the graphon pooling method outperforms the selection GNN, but achieves lower accuracy than graph coarsening. In the case of the polynomial graphon, graphon pooling outperforms both methods with a significantly smaller variance. For all graphons, the smaller variances achieved by graphon pooling can be explained  partly by the fact that the graphon pooling technique does not require choosing which nodes to keep, eliminating node selection bias. This could also be the reason why graph pooling outperforms the selection GNN in all scenarios. Finally, we note that, unlike the bilinear and polynomial graphons, the exponential graphon with parameter $\beta = 2.3$ is not smooth; hence, integrating it over a coarser grid might result in graphs that do not resemble much the original graph. This explains the poorer performance of graphon pooling with respect to graph coarsening in the exponential case.

In our second analysis, we investigate the effect of the layer-wise dimensionality reduction ratios $N_1/N$ (layer 1) and $N_2/N_1$ (layer 2) on GNN performance. We fix the polynomial graphon $W(x,y) = 0.5(x^2+y^2)$ and vary the number of nodes $N$ and the numbers of selected nodes $N_1$ and $N_2$ as indicated in the rows of Table \ref{tb:ratios}. The reported accuracies are averaged over 10 dataset realizations. 
In the first three rows, we fix the reduction ratio of the first layer at $N/N_1 = 2$ and vary $N_1/N_2$. We observe that graphon pooling is outperformed by both graph coarsening and the selection GNN.  
In the following rows, we fix $N_2/N_1$ and only modify $N_1/N$. In this scenario, graphon pooling outperforms both methods, and we observe that as $N/N_1$ increases the gaps in accuracy also increase. In the last row in particular, graphon pooling achieves almost 70\% accuracy by only keeping 5\% of the original nodes. We conclude that graphon pooling is the most reliable pooling strategy when the graph size is large but GNN layer size is restrained. This is the case, for instance, of systems with limited memory or high speed requirements for training and inference. 

\begin{table}[t]
\centering
\begin{tabular}{l|ccc} \hline
 					& \multicolumn{3}{c}{$W(x,y)$} \\
Architecture   		& $\exp (-\beta(x-y)^2)$  & $xy$ & $0.5(x^2 + y^2)$  \\ \hline
Graphon	pooling		& $46.4 \pm 18.4$ & $89.4 \pm 8.0$ & $91.2 \pm 5.3$ \\ 
Graph coarsening	& $66.3 \pm 27.2$ & $99.7 \pm 24.5$ & $87.5 \pm 16.8$  \\
Selection GNN		& $37.8 \pm 14.3$ & $73.0 \pm 31.0$ & $69.6 \pm 20.7$  \\ \hline
\end{tabular}
\caption{Source localization test accuracy (\%) achieved by graphon pooling, graph coarsening and selection GNN on $200$-node graphs obtained from exponential, bilinear and polynomial graphons.}
\label{tb:graphons}
\end{table}

\begin{table}[t]
\scriptsize
\centering
\begin{tabular}{c|c|ccc} \hline
$[N,N_1,N_2]$   	& $\left[\dfrac{N}{N_1},\dfrac{N_1}{N_2}\right]$  & Graphon & Coarsening & Selection  \\ \hline
$[100,50,10]$		& $[2,5]$ & $96.8 \pm 3.8$ & $99.5 \pm 0.4$  & $97.8 \pm 3.9$ \\
$[200,100,10]$		& $[2,10]$ & $70.4 \pm 16.6$ & $87.5 \pm 26.2$  & $78.5 \pm 23.1$ \\ 
$[400,200,10]$		& $[2,20]$ & $61.7 \pm 10.6$ & $70.9 \pm 25.7$  & $63.6 \pm 21.2$ \\ \hline
$[100,20,10]$		& $[5,2]$ & $99.0\pm 0.9$ & $98.2 \pm 2.3$  & $96.3 \pm 6.5$ \\ 
$[200,20,10]$		& $[10,2]$ & $91.2 \pm 5.3$ & $87.5 \pm 16.8$  & $69.55 \pm 20.7$ \\  
$[400,20,10]$		& $[20,2]$ & $69.5 \pm 9.8$ & $47.7 \pm 3.8$  & $52.2 \pm 30.1$ \\  \hline

\end{tabular}
\caption{Source localization test accuracy (\%) achieved by graphon pooling, graph coarsening and selection GNN on graphs obtained from $W(x,y) = 0.5(x^2 + y^2)$, for different values of $N$, $N_1$ and $N_2$.}
\label{tb:ratios}
\end{table} 

\subsection{Movie rating prediction} \label{sbs:movie}

For this experiment, we build a user similarity network from the MovieLens 100k dataset \cite{harper16-movielens}, which consists of 100,000 ratings given by $U=943$ users to $M=1682$ movies. This network is built by computing Pearson correlations between the ratings that pairs of users have given to the same movies \cite{weiyu18-movie, ruiz19-inv} and keeping the 50 nearest neighbors to each user. The full network, with all $U$ users, can then be used to define a step function graphon with $U \times U$ blocks.

The graph data consists of the movies' rating vectors, where the $u$th element of the vector corresponding to movie $m$ is the rating---between $1$ and $5$---given by user $u$ to movie $m$, or $0$ if she has not yet rated this movie. Given a movie's incomplete rating vector, our objective is to predict the rating given by, say, user $u=1$ to any movie $m$. We do this by feeding GNN models with movie vectors where the ratings given by user 1 have been zeroed out and then computing the mean squared error (MSE) loss between the real and the predicted ratings.

To assess the advantages of graphon pooling in this setting, we train a GNN with graphon pooling, one with graph coarsening and a selection GNN to predict the ratings given by user 1. All architectures contain 2 layers, with $F_1=32$ features in the first layer, $F_2=8$ in the second, and $K_1=K_2=5$ filter taps in both. We use 90\% of the movies rated by user 1 for training (out of which 10 \% are used for validation) and 10\% for testing, and train the GNNs over $40$ epochs with learning rate $10^{-3}$ and batch size $5$. 

Two scenarios are analyzed. In the first, we select $N_1 = 100$ and $N_2 = 10$ nodes in the first and second layers respectively, and $N_1 = 50$ and $N_2 = 10$ in the second. The average prediction RMSEs over 10 different train-test splits are presented in Table \ref{tb:movies}. For $[N_1, N_2] = [100,10]$, the GNN with graph coarsening achieves lower test RMSE than the one with graphon pooling; but when $N_1$ is reduced to $50$, graphon pooling outperforms both the graph coarsening and the selection GNNs, corroborating the findings of subsection \ref{sbs:sourceloc}. In the second scenario, we also look at the evolution of the training and validation losses of the three models, whose average curves are graphed in Figure \ref{fig:loss}. Note that, while the graphon pooling GNN only slightly overfits the training set, the graph coarsening GNN overfits it significantly. As for the selection GNN, the large error bars reflect the selection bias associated with the sets of nodes that are kept in each run.

\begin{table}[t]
\centering
\begin{tabular}{l|c|c} \hline
Architecture    & $[N_1,N_2] = [100,10]$  & $ [N_1,N_2] = [50,10]$  \\ \hline
Graphon	pooling		& $1.0987 \pm 0.1196$ & $1.1783 \pm 0.0977$  \\ 
Graph coarsening		& $0.9791 \pm 0.1287$ & $1.1932 \pm 0.1747$  \\
Selection GNN		& $1.0996 \pm 0.1264$ & $1.2871 \pm 0.3868$  \\ \hline
\end{tabular}
\caption{Prediction RMSE for user 1's ratings to movies in the test set. Average over 10 train-test splits. The number of nodes is $N=943$, and $[N_1,N_2]$ stands for the number of selected nodes in the first and second layers of the GNN.}
\label{tb:movies}
\end{table}

%% file: sec_conclusions.tex

In this work we introduced graphons as a tool to perform pooling in a GNN, obtaining the graphs in each layer of the GNN from the graphon by sampling. Graphon pooling preserves the structural and spectral properties of the graph, thus leading to layers where the spectral filtering is consistent, and also eliminates the ambiguity in the mapping of signals between layers.  
The numerical experiments provide clear evidence that this approach outperforms other approaches in the literature when there is a big change in the size of the graph from the first layer to the second one. Additionally, this approach offers considerably less overfitting in comparison with the other approaches, and reduces errors produced by the selection bias associated with choosing  which nodes to keep when pooling is done through zero padding. 

Our approach also opens the door for the design of GNN architectures in which is possible to handle the uncertainty in the properties of the graphs used in each layer as a graphon itself can model a family of random graphs.